\documentclass{article}

\usepackage[english]{babel}

\usepackage[letterpaper,top=2cm,bottom=2cm,left=3cm,right=3cm,marginparwidth=1.75cm]{geometry}

\usepackage{amsmath}
\usepackage{amssymb}  
\usepackage{graphicx}
\usepackage[colorlinks=true, allcolors=blue]{hyperref}
\usepackage{bm}
\usepackage{url}
\usepackage{booktabs}
\usepackage{bbding}
\usepackage{pifont}
\usepackage{wasysym}
\usepackage{utfsym}
\usepackage{fontawesome}
\usepackage{lineno}
\usepackage[final]{changes}
\usepackage{color}
\usepackage{caption} 
\usepackage{float} 

\title{RMAFF-PSN: A Residual Multi-Scale Attention Feature Fusion Photometric Stereo Network}
\author{Kai Luo, Yakun Ju $^{}$*, Lin Qi $^{}$*, Kaixuan Wang and Junyu Dong}

\begin{document}
\maketitle
\makeatletter
\g@addto@macro\@floatboxreset\centering
\makeatother

\begin{abstract}
Predicting accurate normal maps of objects from two-dimensional images in regions of complex structure and spatial material variations is challenging using photometric stereo methods due to the influence of surface reflection properties caused by variations in object geometry and surface materials. To address this issue, we propose a photometric stereo network called a RMAFF-PSN that uses residual multiscale attentional feature fusion to handle the ``difficult'' regions of the object. Unlike previous approaches that only use stacked convolutional layers to extract deep features from the input image, our method integrates feature information from different resolution stages and scales of the image. This approach preserves more physical information, such as texture and geometry of the object in complex regions, through shallow-deep stage feature extraction, double branching enhancement, and attention optimization. To test the network structure under real-world conditions, we propose a new real dataset called Simple PS data, which contains multiple objects with varying structures and materials. Experimental results on a publicly available benchmark dataset demonstrate that our method outperforms most existing calibrated photometric stereo methods for the same number of input images, especially in the case of highly non-convex object structures. Our method also obtains good results under sparse lighting conditions. 
\end{abstract}

\section{Introduction}

Since the creation of the first photometric stereo (PS) algorithm by Woodham \cite{r1} under the Lambert hypothesis, acquiring images with varying light directions using linear response cameras and utilizing the PS algorithm to obtain accurate normal maps of objects have been a focus of researchers, especially in areas where the object's structure and texture have undergone changes~\cite{r2,r3,r4}.
Compared to traditional methods, deep neural networks have the capability to imitate intricate global lighting effects that cannot be represented by previous mathematical formulas, resulting in significantly enhanced accuracy of the results. However, unlike other computer vision tasks that often have a fixed input size or sequence, photometric stereo networks require handling an unknown order problem, making it difficult for Convolutional Neural Networks (CNNs) and Recurrent Neural Networks (RNNs) to handle due to their limited flexibility.

\par
To address the order-agnostic nature of the photometric stereo task and produce an accurate normal map in a complex region, previous studies have proposed several approaches. These include using fused feature maps from max-pooling operations or fixed-size observation maps as inputs to neural networks, increasing the network's depth to improve feature representation, and generating virtual training datasets with varying complexity levels to enhance the network's fitting ability~\cite{r5,r6,r7}. In summary, existing approaches primarily focus on providing various solutions to intricate global illumination issues, including but not limited to shadows and specular highlights. While these approaches have improved the accuracy of normal map prediction, they still face difficulties in dealing with ``difficult'' regions, such as the structure changing rapidly and the material changes, due to the lack of integration of different stages' features 
 of the input image.
\par
Recently, research has shown that low-level vision tasks, including photometric stereo, occur at multiple scales in natural scenes and require careful consideration of scale information~\cite{r8,r9}. On the contrary, many existing methods rely on complex network structures and ignore the fact that scale information changes the focus of one's observation of an image. For example, high-resolution images input at the initial stage of the network can extract rich texture details that help to focus on complex structural regions, whereas low-resolution images at deeper levels tend to emphasize contour information, which facilitates the identification of surfaces with spatially varying materials. Therefore, extracting multi-sale image features from different stages is crucial to solving the problem of fuzzy details in normal~graphs.

In addition to extracting multi-scale features,  since the high-dimensional feature map of the image contains many characteristics of the object surface, attention mechanisms~\cite{r10,r11} can also be used to adaptively adjust feature weights while preserving all extracted features, enabling the network to focus on specific local areas such as material changes or areas affected by global light for optimized feature extraction. Such attention mechanisms have been shown to improve the expressive power of the network and enable it to treat different areas of the image differently, thereby improving the overall performance of the photometric stereo network. Combined with this theory, spatial attention assigns higher weight parameters to areas that require greater attention, such as surface structure or material changes, while the channel attention enhances the weight of channels related to normal regression and suppresses irrelevant channels,  for example, light intensity or roughness-related channels, allowing the network to retain the most useful information with respect to predicting the normal vector of the ``difficult'' region. 
\par
\begin{figure}[htb]
   \includegraphics[width=11.5 cm]{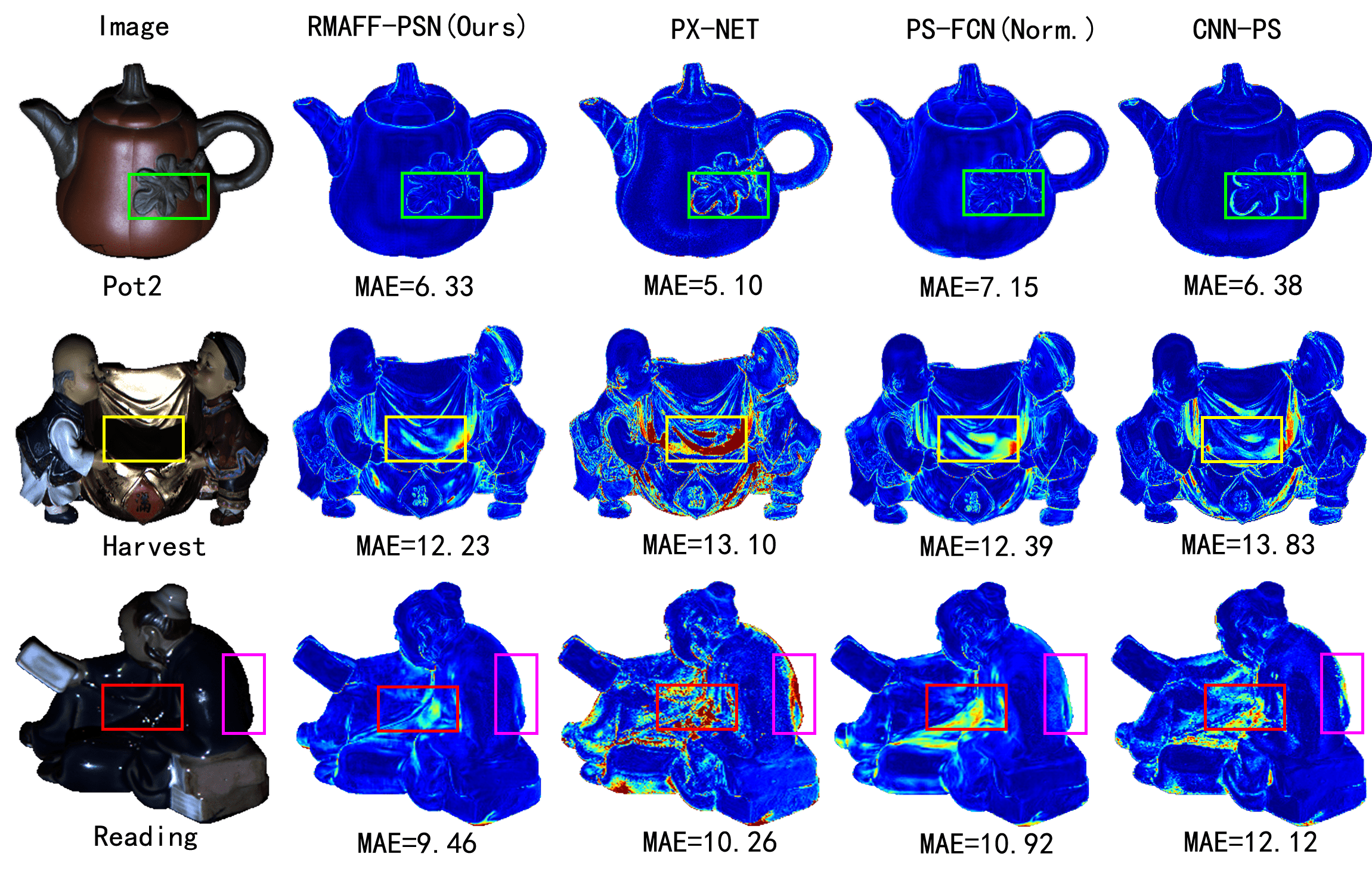}
    \caption{Visualization of structurally complex areas using error maps. The number represents the mean angular error (MAE) of the object. We use green boxes to indicate the material change area, yellow boxes to indicate additional shadows, red boxes to indicate complex areas, and magenta boxes to indicate diffuse reflections. Through our proposed method, we have observed that the accuracy of the restoration process in these areas is significantly improved, as can be seen from the error maps. }
    \label{fig:Key_Figures}
\end{figure}

This paper proposes a novel photometric stereo network, the residual multi-scale attention feature fusion photometric stereo network (RMAFF-PSN), which effectively integrates multi-scale feature information from both shallow and deep stages using attention-weighted fusion. As shown in the boxes in Figure \ref{fig:Key_Figures}, the RMAFF-PSN can effectively deal with the challenges posed by both retaining salient details of object material variations and structurally complex regions during network propagation. To achieve this, we constructed a residual multi-scale attention feature fusion module inspired by previous work on residual multi-scale network structure \cite{r12,r13,r14}. To process the input image, we extract feature information separately from its high- and low-resolution versions and then stitch together the shallow and deep features. The residual multi-scale attention feature fusion module leverages multiple view fields across various scales to enhance multi-scale feature information, in a similar manner to residual connections. Subsequently, the attention mechanism optimizes the feature maps along the channel and spatial dimensions, followed by feature combination with the shallow-deep feature information to complete the steps of feature extraction, enhancement, and optimization. This approach enables the network to effectively capture characteristic information that best describes the different areas of an object's surface, so as to reconstruct the rich and accurate surface normal map.

\par
The results show that the performance of our network is significantly improved compared with the previous method.\ We conducted quantitative experiments on the DiLiGenT benchmark dataset to demonstrate the effectiveness of our approach in dealing with ``difficult'' areas. Additionally, ablation experiments and testing on other real datasets show that the RMAFF-PSN is scalable for photometric stereo tasks.
\par
To sum up, our main contributions are as follows:
\begin{itemize}
    \item For the PS task, we propose a novel residual multi-scale attention feature fusion photometric stereo network (RMAFF-PSN). This model is designed to achieve intensive and precise restoration of 3D shapes, particularly in areas of the object surface that have undergone significant changes(e.g., changes in material).  We believe this contribution provides an innovative approach to restoring complex structures with high accuracy. 
    \item From the scale and attention perspectives, we have designed a residual multi-scale attention feature fusion module. This approach leads to a more effective optimization method for normal correlation feature extraction in the photometric stereo task.
    \item The simple PS data were taken from the number of light sources and reality. This dataset provides a reference for testing the performance of photometric stereo networks in difficult material and geometrically scenarios.
\end{itemize}

\section{Related Work}
Deep learning has demonstrated remarkable efficacy in the field of optical image processing. Santo~\cite{r15} was the first to apply deep learning methods to photometric stereo, achieving prediction accuracies that significantly surpassed those of traditional algorithms. Currently, the prevailing view is that deep learning approaches are predominantly used to tackle photometric stereo tasks using per-pixel and all-pixel methods~\cite{r16,r17}.
\par
In the per-pixel approach, the authors propose using observation maps to solve photometric stereo problems. By mapping pixel values of a given point under all illuminations onto a two-dimensional plane with a fixed size,  the unstructured input is converted into a structured observation map. A convolutional neural network is then trained on this map to perform regressions. The observation map reflects the distribution of outliers over the spatial domain and more accurately captures the intensity variations of pixel points between different photometric images under different illuminations at the same location. 
\par
The all-pixel approach uses pooling operations to aggregate images in all lighting directions, also producing a structured input. This approach is based on the properties of a fully convolutional network and allows for training and testing on input images of any size, making it highly flexible. By using feature extraction and feature fusion operations, this approach effectively explores the variability of internal image regions.
\par
To summarize, these two deep learning-based methods follow a three-step process that includes: (a) feature extraction, (b) feature aggregation, and (c) normal prediction. When given a set of randomly ordered images and their corresponding lights, the feature extraction and aggregation steps can be expressed as follows:
\begin{equation}
    \bm{f_i} = Combin\{ {\bm{x_{1,i}},...\bm{x_{m,i}}}\} 
\end{equation}
Here, 
 $\bm{f_i}$ represents the feature vector of the $ith$ pixel, $\bm{x_{j,i}}$ is the feature value of the $ith$ pixel in the $jth$ light direction. $Combin$ denotes the aggregation of features from the same pixel captured under different lighting conditions. 
\par
The main difference between the per-pixel and all-pixel methods lies in the way feature extraction and aggregation are performed. Therefore, for a given aggregated feature $\bm{f_{i}}$, we can write the normal prediction step as $\bm{n_i} = \Omega (\bm{f_{i}})$ or $\bm{n} = \Phi ({\bm{f_{i}}},...,{\bm{f_{hw}}})$, where $\Omega $ is the per-pixel normal regressor, and $\Phi$ is the all-pixel normal regressor.
\par
Both the per-pixel and all-pixel methods have their respective drawbacks. The former ignores internal image information and is limited by the size setting of the observation map, while the latter overuses 3 $\times$ 3 convolutions and may miss important details. In response to these issues, Yao \cite{r18} proposes using SGC filters to extract feature information from topologically adjacent points, and combines these two methods in a sequential manner. Ju~\cite{r19} adopts a self-supervised approach that learns the attention-weighted loss for each pixel point and introduces penalties for different surface areas, which can retain more detailed gradient information.
\par
However, previous advanced methods are concerned with the design of a deep normal regression network or loss function, ignoring the influence of the relevant normal vector features obtained in the image feature extraction stage on the neural network. Along with this idea, we propose an attention fusion framework that considers the hidden information of image features at different stages and scales. Our framework naturally enhances the characteristics of high-frequency complex regions to improve the accuracy of photometric stereoscopic calculations.

\subsection*{Preliminaries\\}
Before introducing our proposed method, we provide a brief overview of the basic setup and principles of photometric stereo. In calibrated photometric stereo, the goal is to recover the surface normal map of an object from an image captured by a known fixed camera with known illumination directions. Assuming an orthogonal projection camera with a linear radiometric counterpart and a directional light source from the upper hemisphere, the viewing direction($\bm{v}$ = $[0,0,1]^{T}$) is parallel to the z-axis and points towards the origin of the world coordinate system. When global illumination effects, such as inter-reflection and ambient illumination, are absent, the imaging model can be expressed as~follows$:$
\begin{equation}
    I_{j}=s_{j} \rho (\bm{n} ,\bm{l_{j}},\bm{v})max(\bm{n}^{T} \bm{l_{j}},0)+\mu _{j} 
\end{equation}
\par
\begin{figure}[htb]
    \includegraphics[width=12.5 cm]{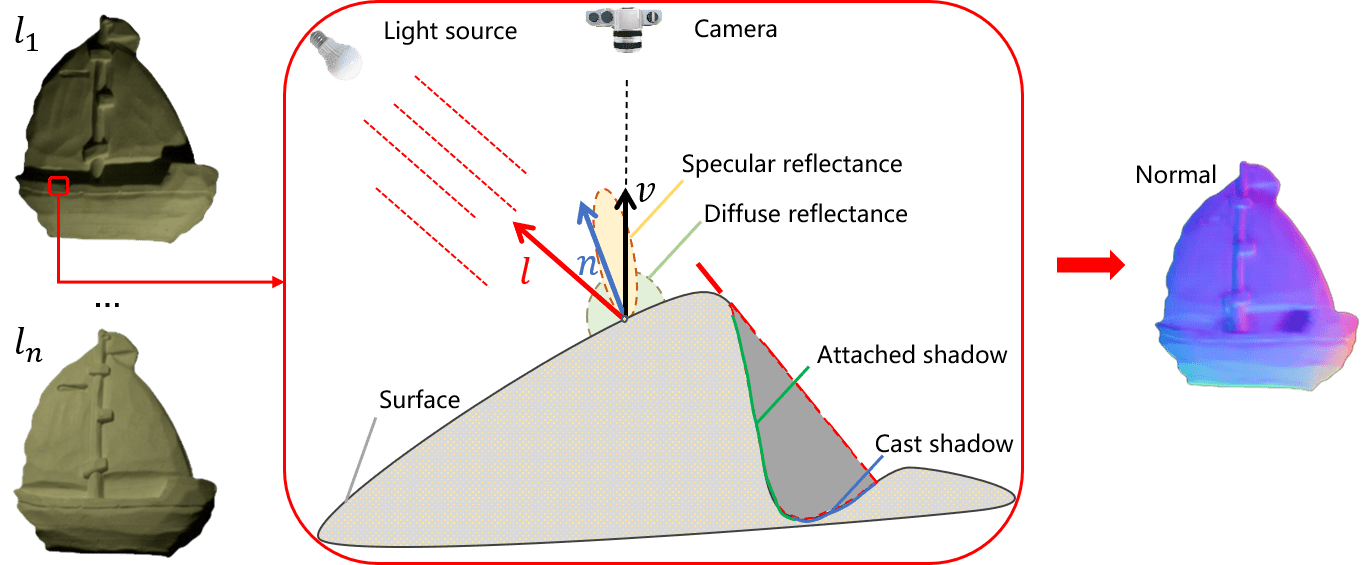}
    \caption{An example of some images with different light directions. In the red box, we illustrate a situation where an object surface point with a normal vector $\bm{n}$ is illuminated by an infinitely distant point light source in a direction $\bm{l}$, and is observed by a camera in a view direction $\bm{v}$. When $\bm{n}^{T} \bm{l_{j}}<0$, an additional shadows occur, and a cast shadows appear when the light is occluded by the object.}
    \label{fig:WhatPS}
\end{figure}

Here  $I_{j}$ represents the image pixel intensity in the $jth$ illumination direction. $\rho$ is the bidirectional reflectance distribution function (BRDF, an important formula in the field of optics and graphics used to describe how light is reflected from a given direction of incident light and outgoing light at a surface~\cite{r20}) and $ max(\bm{n}^{T} \bm{l_{j}},0)$ represents an attached shadow, and $\mu _{j} $ represents the noise of the camera and the environment.  We assume that the pixel intensities of an image are normalized by the corresponding lighting intensities, so $\bm{l}$ can be viewed as a unit vector.
The challenge of dealing with non-Lambertian surfaces limits the applicability of some photometric stereo methods that are suitable only for Lambertian surfaces, such as the least squares method. To address this issue, deep learning methods have emerged as a powerful tool for fitting surfaces of objects that cannot be expressed by mathematical formulas.\ However, even with the impressive fitting ability of deep learning methods, reconstruction results can still be fuzzy and have large angle errors due to the shape or material of some areas. As shown in Figure  \ref{fig:WhatPS}, complex scenes with intricate structures can be particularly challenging. Fortunately, this paper proposes a novel approach to address these ``difficult'' regions using multi-scale attention feature fusion, which is both simple and effective.

\section{Methods}
\begin{figure}[htb]
    \includegraphics[width=12.5 cm]{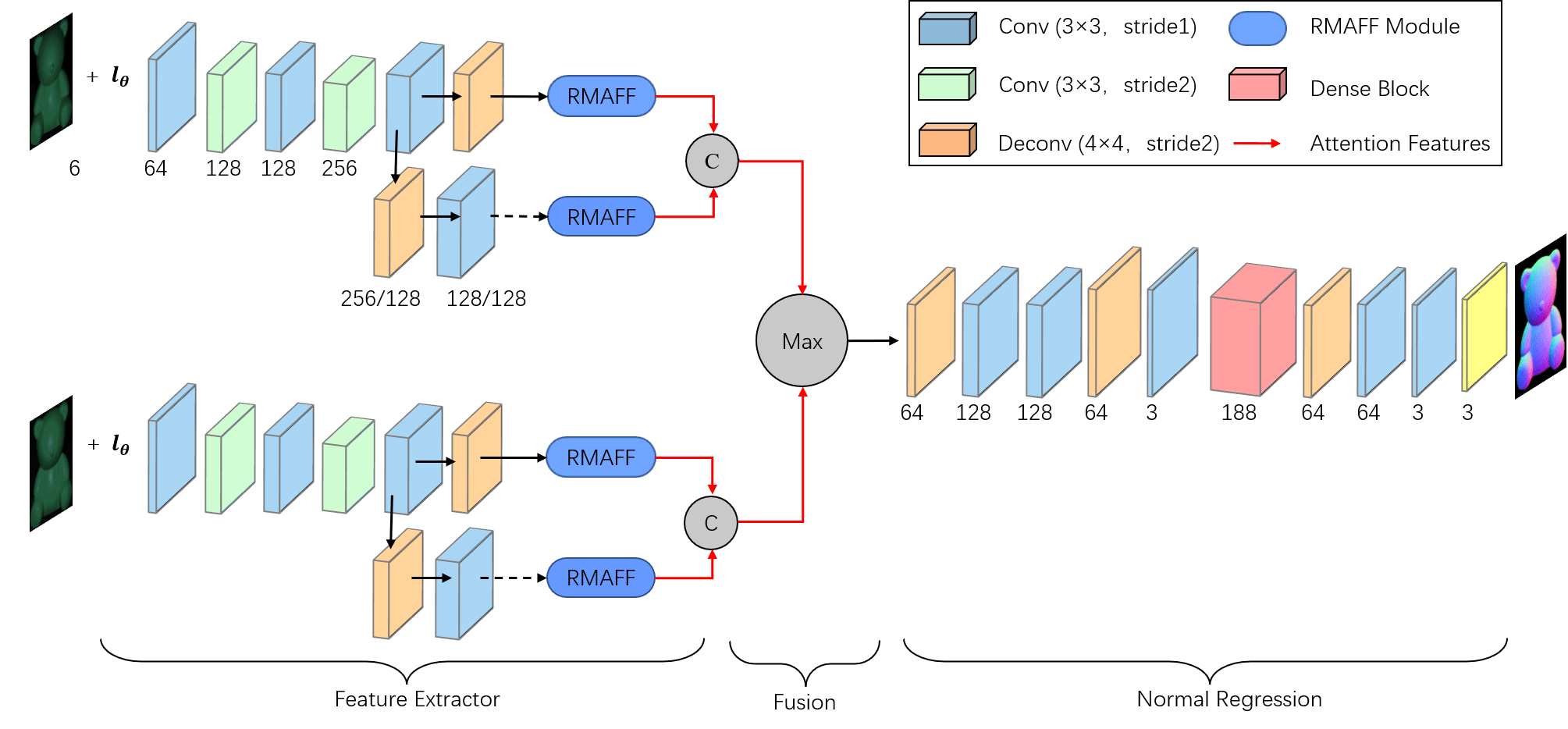}
    \caption{RMAFF-PSN network architecture. The number underneath each layer refers to the number of the channel that is used in the convolution.}
    \label{fig:Network_Structure_3}
\end{figure}

This paper introduces a novel calibrated photometric stereo network that integrates residual multi-scale attention feature fusion. The proposed model structure is illustrated in Figure  \ref{fig:Network_Structure_3}. To preserve effective normal-correlated feature information in the feature extraction stage and avoid feature loss due to redundant convolution operations,  we employ the high-resolution image from the shallow stage and the low-resolution image from the deep stage to retain the texture and contour features of the object surface, respectively. By combining these features, we achieve more accurate normal map prediction in regions with complex structures and spatially varying materials. Ablation experiments are conducted to comprehensively evaluate the network performance by utilizing unbalanced feature information across the shallow and deep layers, to 
further enhance the multi-scale information, and focus more attention on regions with rich structural and material feature information and less attention on ordinary diffuse regions. We specially design a residual multi-scale feature attention fusion module, abbreviated as the RMAFF module, to enhance and optimize the multi-scale information from the shallow and deep stages of the stitched~image.
\par
The order and number of the photometric stereo tasks can vary in the feature fusion stage, causing uncertainty in the image input order. As opposed to conventional CNNs, images cannot be sequentially operated on in this stage. To address this challenge, pooling operations are utilized to fuse different image features in an order-agnostic manner. Specifically, the max-pooling operation is utilized to acquire the most expressive feature information in the image along the same channel dimension of the image features under different illumination directions. By aggregating feature maps through max-pooling, our proposed method can effectively capture the most salient features across different images, resulting in improved accuracy of photometric stereo reconstruction, especially in high-frequency regions of the object surface.
\par
In the normal regression part, to prevent overfitting caused by a deep network, we added a dense-block structure~\cite{r21}. This structure enhances feature propagation and reuses features from low-dimensional inputs to improve the accuracy of the network in predicting the normal image pixels while retaining the features of shallow local regions. In addition, we used batch normalization between each layer of the convolutional neural network to adjust the weights of the neurons into a standard normal distribution regularization layer. Our network architecture consists of six convolutional layers, two downsampling layers, five upsampling layers, two residual multi-scale attention feature fusion modules, one max-pooling layer, and one dense-block module. Furthermore, we utilized L2-Norm layers to normalize the surface normal vectors.

\subsection*{Residual Multi-Scale Attention Feature Fusion Module\\} 

While residual blocks have proven to be successful in capturing multi-scale features, they may not fully capture all multi-scale features with the use of solely 3 $\times$ 3 convolutional kernels, which can lead to feature loss. To overcome this limitation, we propose the use of a Residual Multi-scale Attention Feature Fusion (RMAFF). The RMAFF module incorporates the attention mechanism and multi-scale feature representation to more effectively capture regional features in images, enhancing the ability of the network to learn more comprehensive and representative normal vector-related features.
\par
\begin{figure}[htb]

    \includegraphics[width=8.5 cm]{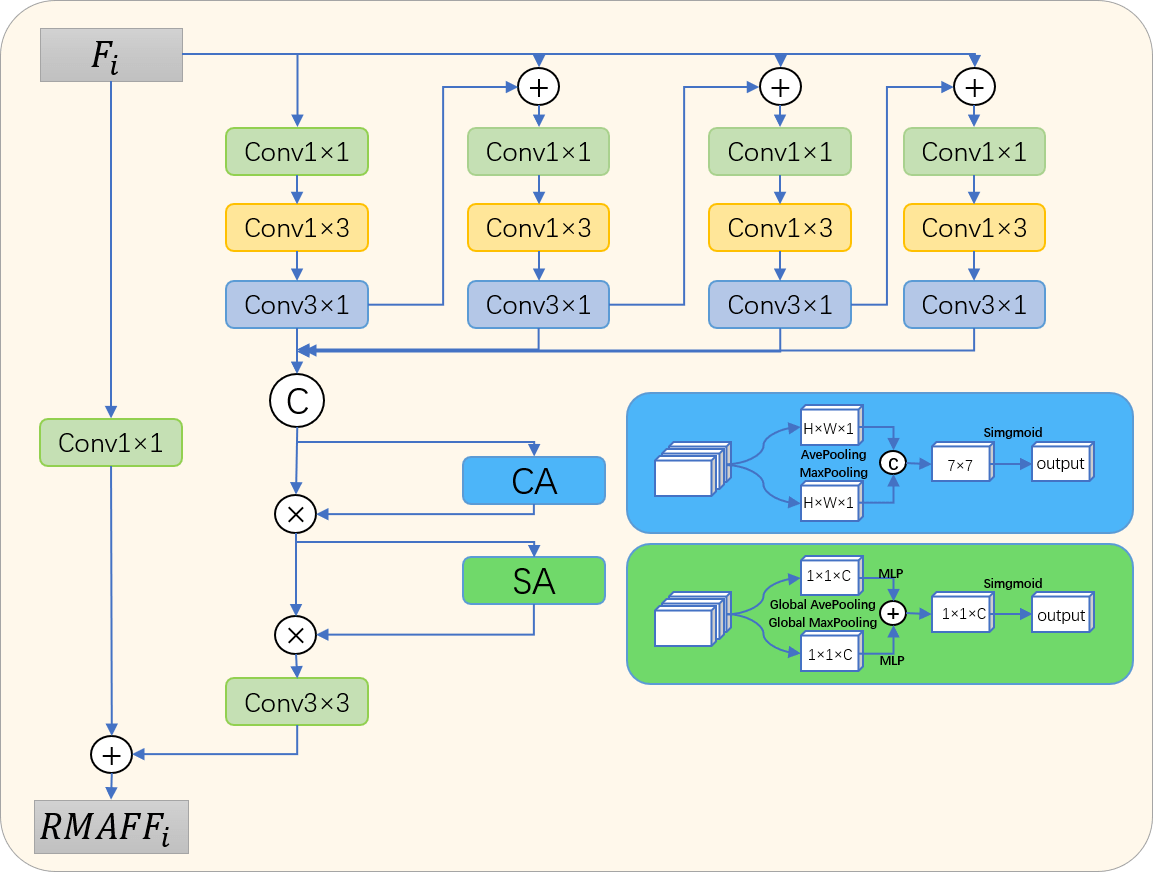}
    \caption{Structure diagram of RMAFF module. It uses residual-like blocks to expand the field view while adaptively adding attention weights to feature information.}
    \label{fig:RMAFF_Module}
\end{figure}

As depicted in Figure  \ref{fig:RMAFF_Module}, the RMAFF module utilizes a residual-like block to incrementally enlarge the network's receptive field while improving the extracted features through the use of an asymmetric convolution kernel. This technique enables the capturing of comprehensive feature information, while channel attention and spatial attention mechanisms are employed to guide the network in optimizing high-frequency regions of the image from both global and local perspectives.

\par
More specifically, the module comprises four branches that are utilized to capture different features of the input feature map $\bm{F_{i}}$. Each branch begins with a 1 $\times$ 1 convolution operation to exchange channel information, followed by 1 $\times$ 3 and 3 $\times$ 1 asymmetric convolution operations to highlight local key features in both the horizontal and vertical directions. The output of each branch is added to the input of the next branch to integrate multi-scale features. This operation can be mathematically formulated as follows$:$
\begin{equation}
\bm{Branch_j^i} = \left\{ \begin{array}{l}
AsyConv(\bm{F_{i}}){\rm{ }},{\rm{    }}j = 1{\rm{ }}\\
AsyConv(\bm{F_{i}} \oplus \bm{Branc{h_{j - 1}}}){\rm{ }},{\rm{   }}j = 2,3,4
\end{array} \right.
\label{equation1}
\end{equation}
where $\bm{F_{i}}$ represents the $ith$ feature map, $j$ is the number of branches, and $\bm{Branch}_j^i$ represents the output of the $j$ branch. $\oplus$ represents the pixel-by-pixel summation and $AsyConv( )$ represents the asymmetric convolution layer.
\par
After stitching together the multi-scale augmented features, we apply global average pooling and global maximum pooling operations to obtain two feature vectors.\ These vectors are then used to calculate the channel attention weights using a fully connected layer, which are  multiplied with the feature map to optimize the channel features. The resulting feature map is then subjected to average pooling and maximum pooling operations along the channel dimension.  The outputs of these operations are  concatenated, and the resulting vector is passed through an activation layer to obtain the spatial attention weights. Finally, the original feature map is multiplied by the spatial attention weights to obtain the final output. The overall operation can be formulated as follows$:$
\begin{equation}
    \begin{array}{l}
{\bm{F_{CA}}} = {{\rm{g}}_c}(\bm{F_{{\rm{cat}}}}) \otimes \bm{F_{{\rm{cat}}}}
\end{array}
\label{equation2}
\end{equation}
\vspace{-6pt}
\begin{equation}
    \begin{array}{l}
{\bm{F_{SA}}} = {{\rm{g}}_{\rm{s}}}(\bm{F_{CA}}) \otimes \bm{F_{CA}}
\end{array}
\label{equation3}
\end{equation}
where $\bm{F_{{\rm{cat}}}} \in \mathbb{R}^{c \times h \times w}$ denotes a multi-scale feature with four branches stitched together,
${{\rm{g}}_c}\in\mathbb{R}^{c \times 1 \times 1} $ denotes a $ 1D $ channel attention chart,
${{\rm{g}}_{\rm{s}}}\in\mathbb{R}^{1 \times h \times w} $ denotes a $ 2D $ spatial attention map,
$\otimes$ denotes element-by-element multiplication.
\par
Finally,  a 3 $\times$ 3 convolution operation is performed to adjust the number of channels, and the resulting features are added  to the original features. The RMAFF module can be summarized as follows, without losing generality$:$
\begin{equation}
RMAF{F_k} = Con{v_{1 \times 1}}(\bm{F_i}) \oplus Con{v_{3 \times 3}}(\bm{F_r}(Cat_{j = 1}^4(\bm{Branch_j^i})))
\end{equation}
where $Con{v_{1 \times 1}}$ denotes an 1$ \times $1 convolution operation on the feature map. $\bm{F_r}$ denotes the channel attention operation and the spatial attention operation illustrated in Equations~(\ref{equation2}) and (\ref{equation3}).

\section{Experimental Results}
The model was trained  on a NVIDIA RTX 3090 24G GPU. The initial learning rate was set to 0.001, and was decreased by a factor of two every five epochs.\ The model was trained using 32 batches for 30 epochs, and the best model was selected as the final result. To evaluate the similarity between multiple feature channels, we employed the cosine similarity measure. This approach is a flexible approach and provides a fixed-dimensional scalar for each neighboring point. The cosine similarity loss function is defined as follows$:$
\begin{equation}
l o s s=\frac{1}{h w} \sum_i\left(1-\bm{n_i} \cdot \bm{n_i^{\prime}}\right)
\end{equation}
where $ h $ and $ w $ are image resolutions, $ i $ is the image index, $ \bm{n_i} $ and $\bm{n_i^{\prime}}$ are the true and predicted normals. The more similar they are the closer their product is to one.
\subsection*{Datasets\\}
The insufficient number of datasets is a limiting factor that affects the accuracy of photometric stereo predictions. In the experiments conducted in this study, the training datasets used were the same as those used by most all-pixel methods. The two shape datasets are the Blobby and Sculpture datasets~\cite{r22,r23}, which were rendered using the MERL BRDF dataset~\cite{r24} containing information on 100 different materials. The two datasets contain 25,920 and 59,292 shapes, respectively, and each captured shape has 64 different lighting conditions, resulting in a total of 5.4 million images. The Sculpture dataset is more complex and contains more detailed information. During the training process, we used a ratio of 99:1 for training and validation data.
\par
The DiLiGenT dataset~\cite{r25} consists of 10 real-world objects with complex shapes and different materials, each captured in 96 lighting orientations. Ground truth information is provided for each object in the dataset. To evaluate the effectiveness of our proposed model, we used this dataset for quantitative assessment.
\par
We evaluated the performance of our proposed the model using the mean angular error (MAE) as the evaluation metric. To calculate the MAE, we computed the angular error between the predicted pixel normal value and the ground truth normal value of each pixel. The calculation formula for the angular error between the predicted normal and the ground truth normal is as follows$:$
\begin{equation}
M A E=\frac{1}{h w} \sum_i \arccos \left( \bm{n_i} \cdot \bm{n_i^{\prime}} \right)
\end{equation}
where $ \bm{n_i} $ and $\bm{n_i^{\prime}}$ are the true and predicted normals.
\par
In addition to the DiLiGenT dataset, we also evaluated our proposed model on the Apple and Gourd dataset~\cite{r26}. This dataset contains three different objects, each with approximately 100 images of 646 ${\times}$ 696 resolution. We used this dataset to further verify the performance of our proposed method on real-world objects.

\par
The DiLiGenT${10}^2$ dataset~\cite{r27} was created to overcome the limitations of the DiLiGenT dataset, which lacks diversity in terms of object materials and shapes. This new dataset includes 10 objects, each made of a different material, and generated using an advanced 3D modeling machine that captures more detailed information about their shapes and materials. By controlling the shape and material information, the dataset allows researchers to evaluate the network's ability to handle various material and shape variations.
\par
To showcase the practical usefulness of our proposed model, we acquired a new dataset consisting of six objects illuminated by six fixed light angles, where the zenith angle was fixed at 45° and the directional angles were spaced at 60°. We named this dataset Simple PS data. The images were captured using an IDS industrial camera, which provides high-quality and high-resolution images suitable for testing our model's performance in real-world scenarios.
\par

\begin{figure}[htb]
    \includegraphics[width=12.5 cm]{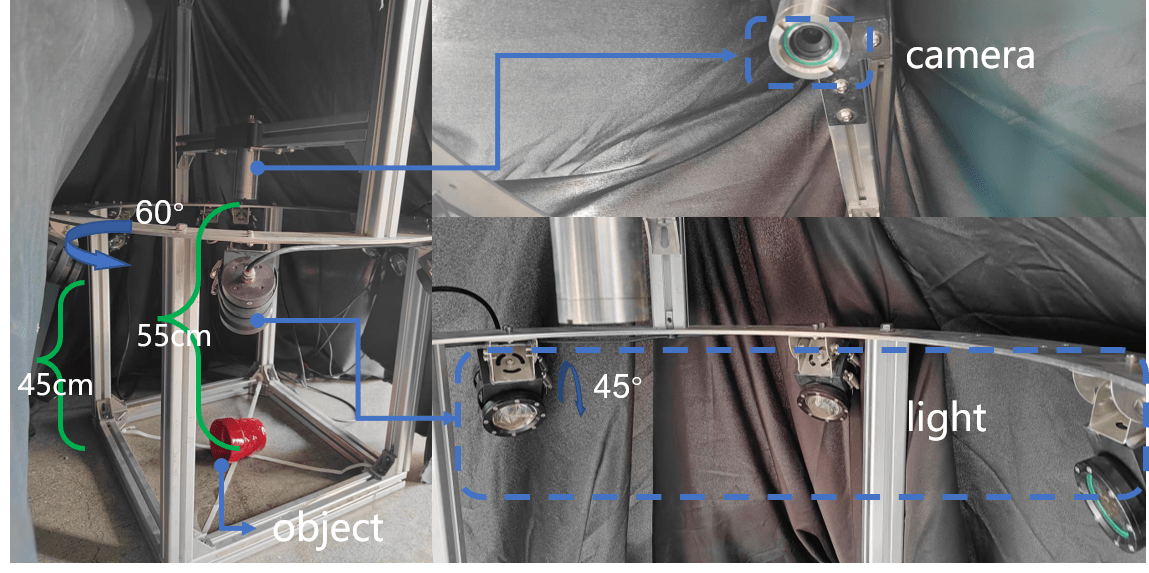}
    \caption{Imaging setup for building the Simple PS data. We built a fabricated shelf and covered it with black cloth to simulate darkroom conditions.\ The camera is fixed at the top of the shelf. Six light sources were installed around the iron ring, and the target object was placed directly below the camera. The blue line shows the detailed device and location information, and the green line shows the height of the device from the ground.}
    \label{fig:HowPhoto}
\end{figure}

In this work, we employed the photometric stereo setup, as shown in Figure \ref{fig:HowPhoto}. The shooting scene was under darkroom conditions, and the object size was 0.25 m
, and the camera was positioned at a fixed height of 0.55 m. The original images were captured at a resolution of 3120 $\times$ 3120 pixels and a bit depth of 24. During the test, the illumination intensity is uniformly set to 1, and since the angle of the light source is fixed, the light source direction matrix can be obtained by using the method of calibration sphere light source calibration. However, due to the substantial effort required to scan the object with a 3D scanner, as well as align and calibrate the captured images, we only used image data to validate the effectiveness of our proposed network model under sparse lighting conditions that occur in real-world scenarios. Our aim is to evaluate the performance of the network model in such practical scenarios.

\par
In light of the fact that the three datasets mentioned above had no corresponding ground truth, we reconstructed normal maps of the objects in these datasets for qualitative visual analysis.

\subsection*{Network Analysis\\}
We illustrate the impact of different network structures on complex object structures, such as the intricate pattern of ``Pot2'', and the folds in the clothing of ``Reading''. In the following sections, we  evaluate the effectiveness of the residual multi-scale attention fusion module, the impact of the normal regression networks with different structures, and the influence of the number and resolution of test images on the accuracy of the resulting object normal maps.

\subsubsection*{Effectiveness of a Residual Multi-Scale Attention Feature Fusion Module\\}
To investigate the effect of the RMAFF module on feature extraction performance, we conducted ablation experiments on the network structure using the same training dataset. Four variants were implemented , namely, ``w/o RMAFF'', ``w/o Attention'', ``Single RMAFF'', and ``RMAFF+AFF''.
\par
``w/o RMAFF'' removes the RMAFF module from the network and only concatenates shallow-deep features. ``Single RMAFF'' employs a single branch RMFE module to extract different levels of features from the deep image. ``w/o Attention'' removes the attention allocation mechanism from the module and directly connects the feature information extracted by the residual multi-scale module. ``RMAFF+AFF'' replaces the concat operation with the AFF structure \cite{r28} for feature stitching. Our approach uses two RMFE modules to extract different features from the shallow and deep layers and then stitches them together using the concat operation.

\par
Table \ref{table1} shows all the network structure variants results, and the corresponding best model results from 30 epochs of training.
From the experimental results, the experimental results of ID(1) are better than those of ID(0) without the feature enhancement and optimization steps.  This is due to the fine-grained global-local multi-scale feature grouping that the features undergo before the pooling and fusion operation.  Additionally, the attention mechanism assigns different attention weights to weaken unimportant features, leading to better results as the network can focus on the important regions of the image.

\par
The MAE for ID(4) is 7.13. The experimental results indicate that  shallow features contain more texture information. The concurrent extraction of both shallow and deep features preserves significant regional characteristics after applying the RMAFF module, which in turn results in better estimations of projected shadow regions and spatially varying material regions.

\par
The  results of ID(2) and ID(3) reveal that the number of attention operations can have an impact on model performance. In particular, ID(3), which redistributes attention weights using the AFF structure to merge the optimized shallow and deep features, performed worse than ID(2). Our hypothesis is that reusing the attention mechanism enhances the most salient information and suppresses the least salient information. However, after fusion and pooling, different channels represent the decomposition of images under varying lighting conditions, and feature maps with significant differences in channel features may have limited representation power in certain shadow-obscured regions. This may explain why the concat operation produced excellent results.\ The distribution of features under one branch reinforces the salient features and weakens the features in the shaded areas, while the opposite may be true for another branch. By max-pooling the features from all branches, the information from each channel can best represent the surface normal distribution of the~object.

\begin{table}[htb]
\caption{RMAFF-PSN ablation experiments with the average angle error value on the real dataset DiLiGenT regarding the accuracy of the RMAFF module feature extraction.\label{tab2}}
\resizebox{\textwidth}{!}{
            \begin{tabular}{clccccccccccc}
\toprule
\textbf{ID}  & \textbf{Method} & \textbf{Ball} & \textbf{Cat} & \textbf{Pot1} & \textbf{Bear} \textsuperscript{\textbf{1}} & \textbf{Pot2} & \textbf{Buddha} & \textbf{Goblet} & \textbf{Reading} & \textbf{Cow} & \textbf{Harvest} & \textbf{Avg.} \\ \midrule
(0) 
 & w/o RMAFF       & 2.66         & 5.82        & 7.71         & 6.69         & 8.06         & 7.75           & 8.96           & 13.33           & 8.26        & 15.31           & 8.45          \\
(1) & Single RMAFF    & 2.48         & 4.68        & 5.99         & 7.04         & 6.91         & 7.55           & 8.72           & 9.72            & 8.68        & 12.47           & 7.42          \\
(2) & w/o Attention   & 2.72         & 4.96        & 6.45         & 7.09         & 7.33         & 7.81           & 8.09           & 8.98            & 6.84        & 12.07           & 7.23         \\
(3) & RMAFF+AFF       & 3.92         & 4.80        & 6.35         & 7.62         & 7.67         & 7.95           & 9.98           & 10.47            & 6.56        & 12.68           & 7.80          \\ 
(4) & Our methods        & 2.18         & 4.64        & 5.52         & 7.53         & 6.33         & 7.71           & 8.26            & 9.46            & 7.50        & 12.23           & 7.13         \\\bottomrule
\end{tabular}}
	\noindent{\footnotesize{\textsuperscript{1} For comparison purposes, all 96 images were used uniformly in our ablation experiments for the ``Bear'' object.}}
 \label{table1}
\end{table}

\par
\begin{figure}[htb]

    \includegraphics[width=12.5 cm]{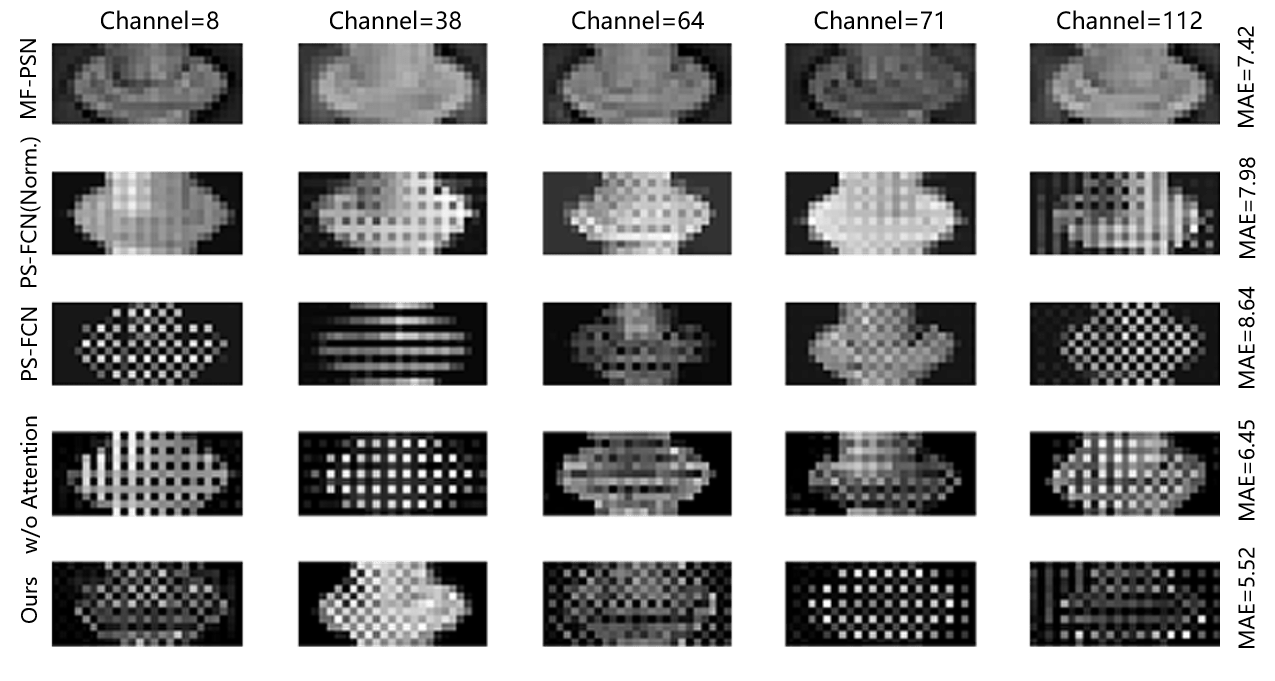}
    \caption{The grayscale map is utilized to visualize certain feature map channels following maximum pooling. We focus on the ``supporting foot'' region of the ``Goblet'' object in the DiLiGenT dataset.}
    \label{fig:Channel_V}
\end{figure}

The comparison between ID(2) and ID(4) demonstrates the effectiveness of the channel and spatial attention in enhancing the normal vector features in the image. We display a selection of feature map channels after maximum pooling in Figure \ref{fig:Channel_V}. Our results indicate that the addition of the attention mechanism in the RMAFF module leads to a significant reduction in the average angle error value. This finding supports the notion that our proposed method can effectively integrate multi-scale features and improve the accuracy of calibrated photometric stereo.

\subsubsection*{The Validity of Normal Regression Network Structures\\}
\begin{figure}[htb]
   
    \includegraphics[width=12.5 cm]{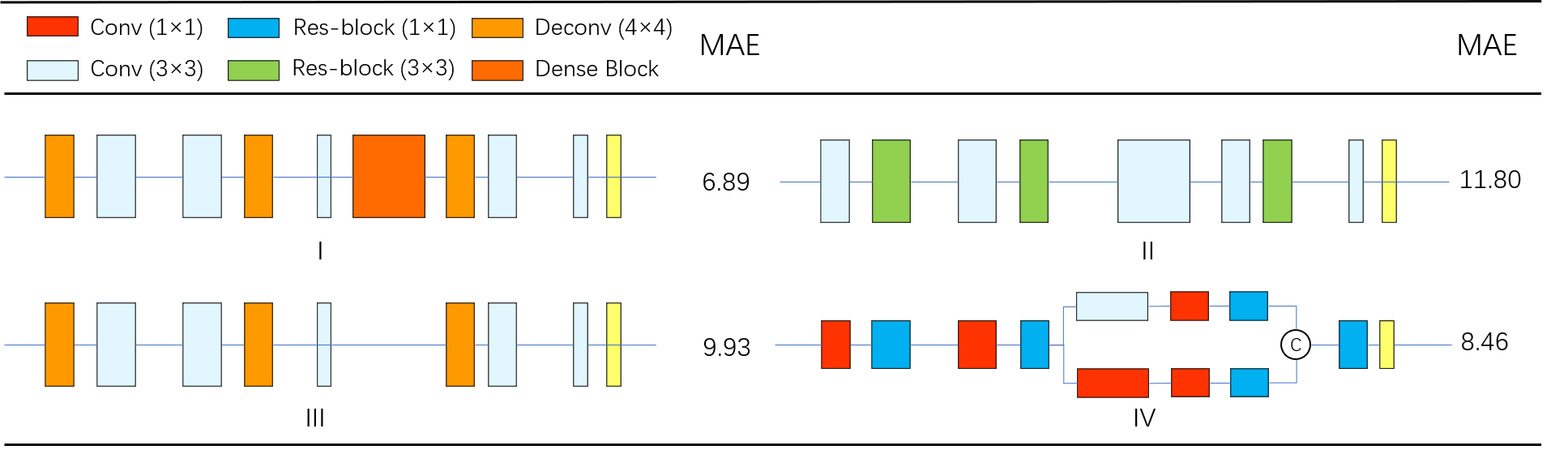}
    \caption{The validity of normal regression network structures.\ To prevent the network from generating redundant information, our normal regression network adds the dense-block module for feature reuse, once again integrating features at different levels.}
    \label{fig:Norm_Network}
\end{figure}

We validated the effectiveness of our normal regression network, as depicted in Figure~\ref{fig:Norm_Network}. Experiments were conducted using all 96 input images of the DiLiGenT dataset. By comparing the results obtained from regression network structures I, II, and III, we found that excessive usage of 3 $\times$ 3 convolutional kernels caused feature smoothing, leading to inferior performance. However, the addition of dense-blocks resulted in superior performance as each hidden layer learned more discriminative features through feature fusion. In the comparison results for I and IV we observed that although the multi-branch design structure utilized spatial information more effectively, using redundant residual connections in the regression part resulted in a significant amount of unnecessary and redundant information, which degraded the accuracy of the pixel normal~prediction.

\subsubsection*{Effect of Different Resolutions and Number of Input Images\\}
Images of different resolutions may contain varying levels of information, with larger images often containing richer features such as geometry and texture~\cite{r29}. Our proposed RMAFF module is designed to fully perceive light and dark variations within an image and extract the most appropriate image features based on those variations. However, due to GPU memory limitations, training models with larger numbers and higher resolutions of input images can significantly increase the training time.
\par
\begin{figure}[htb]

    \includegraphics[width=13.5 cm]{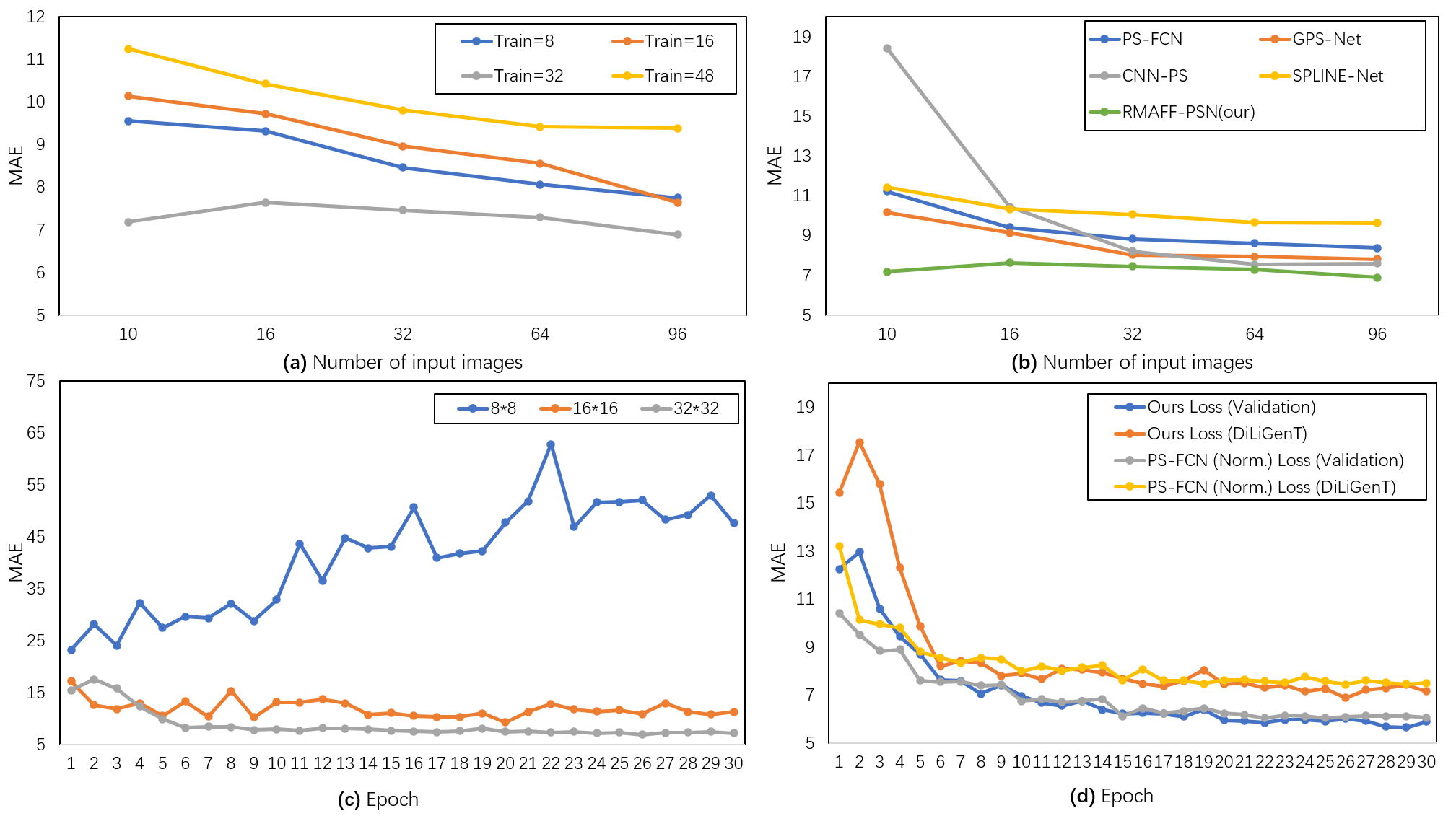}
    \caption{(\textbf{a}) The results of the RMAFF-PSN trained and tested with different numbers of input images. 
(\textbf{b}) The quantitative comparison on the DiLiGenT dataset. The errors for 10 objects are averaged.
(\textbf{c})~The results of the RMAFF-PSN tested with different resolutions of input images.
(\textbf{d}) The comparison of the convergence of our RMAFF-PSN and PS-FCN (Norm.) on the same train dataset and the DiLiGenT benchmark dataset.}
    \label{fig:four_as_one}
\end{figure}

We conducted experiments using the DiLiGenT dataset to examine the effect of input image numbers and resolutions on our model’s training and testing performance. Figure  \ref{fig:four_as_one} illustrates the results obtained from training images with various numbers and resolutions. Our findings indicate that using 32 images at a resolution of 32 $\times $ 32 achieves a good balance between training time and prediction accuracy, as these images contain ample information for the model to learn from. Larger input images tend to offer richer features, but training models with larger input sizes may incur longer training times and lead to more complex model structures. Therefore, it is important to find a balance between input image sizes and model performance in practical applications.

\subsubsection*{Effect of Different Training Datasets\\}
\begin{figure}[htb]

    \includegraphics[width=12.5 cm]{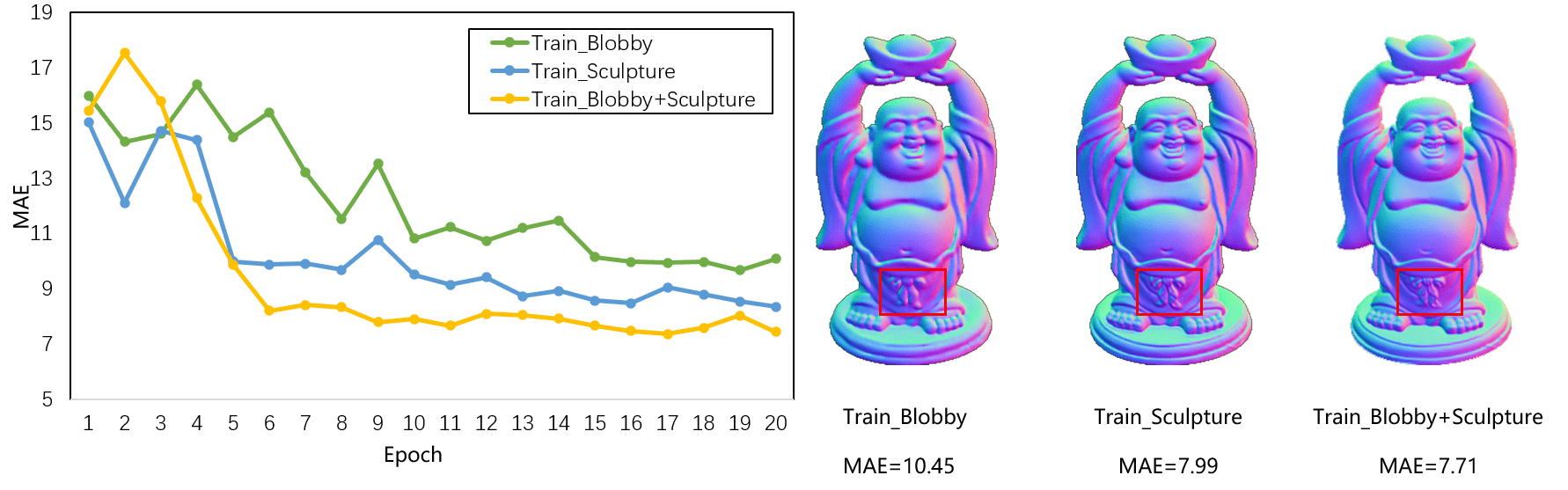}
    \caption{We present the average angle error of our network on the DiLiGenT dataset using different training datasets, along with the reconstructed results of the ``Buddha'' object. The challenging area of the object is denoted by the red box.}
    \label{fig:TrainDatasets}
\end{figure}

In this study, the influence of dataset complexity on the accuracy of the proposed network model is analyzed. It is well known that the accuracy of the network model can be affected by the complexity of the dataset, as even better networks may not necessarily improve model accuracy, a phenomenon known as ``Kolmogorov complexity'' in machine learning~\cite{r30,r31}. To investigate the effect of dataset complexity on network performance, we conducted separate experiments on three different training datasets and reconstructed the normal map of the ``Buddha'' object. It should be noted that everything was kept the same for the network except for the training dataset. The experimental results are shown in Figure \ref{fig:TrainDatasets}. Our analysis indicates that the network's performance in reconstructing surface normals is better when the surface complexity of the training dataset is higher (Train\_Blobby vs. Train\_Sculpture) and the dataset has a larger number of samples (Train\_Blobby vs. Train\_Blobby+Sculpture). Therefore, we conclude that the complexity of the object shape and material is essential in the photometric stereo task, enabling the network to capture more features of the object surface and achieve a more accurate normal estimation during the test phase.

\subsection*{Benchmark Comparison of the DiLiGenT Dataset\\}
\begin{figure}[htb]
    \includegraphics[scale=0.12]{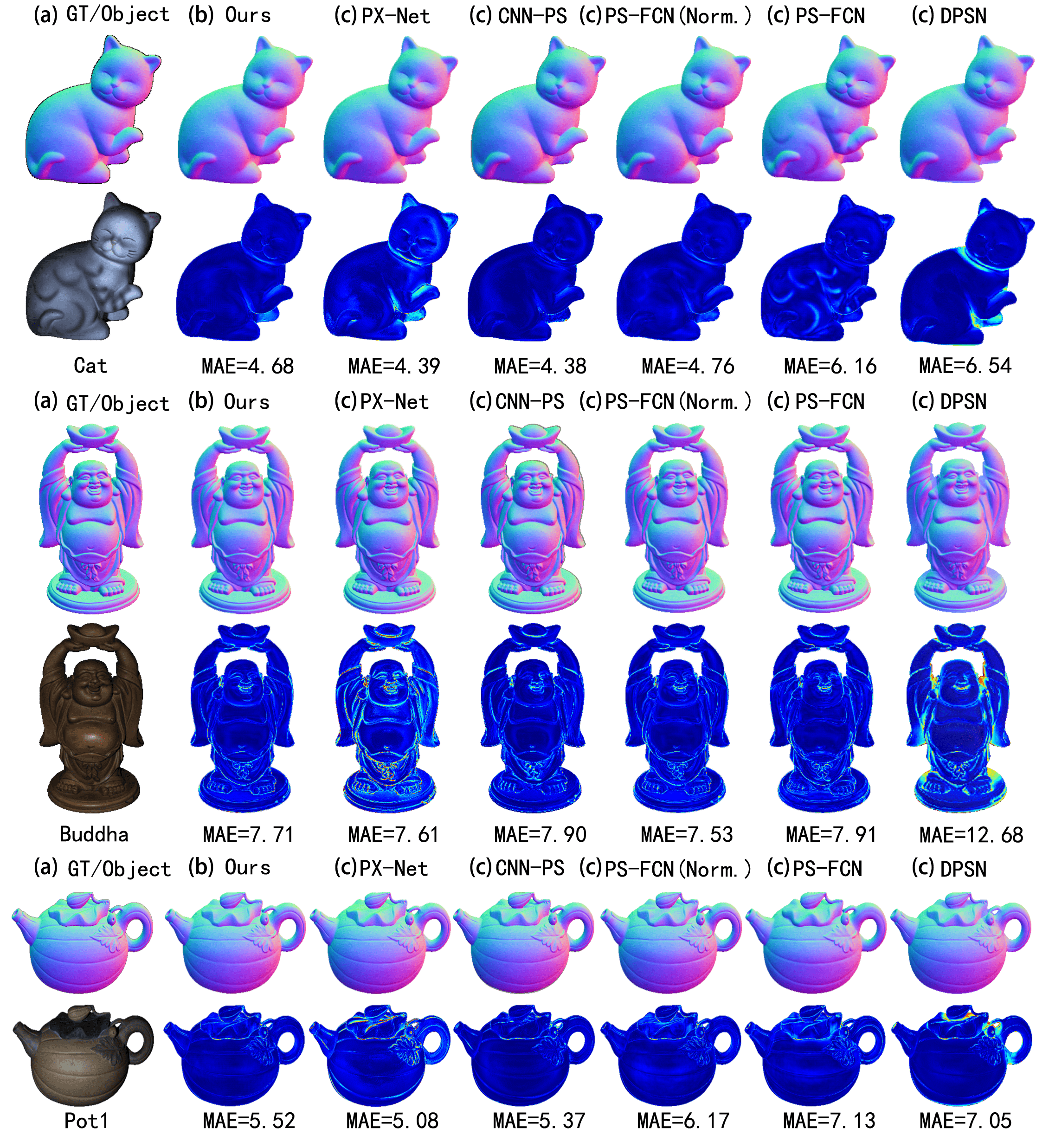}
    \caption{Qualitative results on the DiLiGenT benchmark main dataset. From left to right columns in each scene, we show (\textbf{a 
}) observed images and ground-truth, (\textbf{b}) estimated surface normals and angular error maps by our method, and (\textbf{c}) estimated surface normal and angular error maps by some state-of-the-art methods. The numbers under the error maps indicate their mean angular error in degrees.}
    \label{fig:All_Compare}
\end{figure}

We conducted a comprehensive comparison of the RMAFF-PSN with several commonly used methods, which included linear least squares-based methods (L2~\cite{r1}), four per-pixel methods (CNN-PS~\cite{r6}, LMPS~\cite{r32}, SPLINE-Net~\cite{r33}, PX-Net~\cite{r7}), and four all-pixel methods (IRPS~\cite{r34}, PS-FCN~\cite{r5}, CHR-PSN~\cite{r8}, MF-PSN~\cite{r9}).\ To evaluate the performance of each method, we used the DiLiGenT dataset, which included 96 input images for all methods, except for the ``Bear'' object, which used 76 images due to corruption in the first 20 images. The results of our experiments are presented in Table \ref{table2}.\ Our proposed RMAFF-PSN method outperformed most of the networks, exhibiting higher accuracy than the existing deep learning methods under varying light distributions and image sizes (Figure \ref{fig:four_as_one}b,d). Furthermore, we provide a visual comparison of the predicted normal maps for each object in the DiLiGenT dataset  in Figure  \ref{fig:All_Compare}.

\par
The experimental results presented in Table \ref{table2} unequivocally validate the efficacy  of our proposed method, as evidenced by its impressive average MAE of 6.89 on the DiLiGenT dataset, either ranking first or second among previous methods. Notably, our method delivers particularly exceptional results for objects with intricate structures such as ``Pot2'' and ``Reading'', further underscoring its superiority over existing alternatives in handling complex geometries. This improved performance is mainly attributed to the RMAFF module incorporated into our network structure, which effectively enhances the representation of complex regions and significantly improves the network's ability to recover finer details. For instance, in the ``Harvest'' object, our method delivers significantly finer normals in the pocket region, thereby provding further evidence of the efficacy of the RMAFF module in handling complex geometries.
\renewcommand{\arraystretch}{0.6}
\begin{table}[htb]
\caption{Quantitative comparison of the proposed method with both traditional methods and deep learning methods is conducted on the DiLiGenT benchmark.\label{tab2}}
	\resizebox{\textwidth}{!}{
            \begin{tabular}{lccccccccccc}
\toprule
\textbf{Method} & \textbf{Ball} & \textbf{Cat} & \textbf{Pot1} & \textbf{Bear} & \textbf{Pot2} & \textbf{Buddha} & \textbf{Goblet} & \textbf{Reading} & \textbf{Cow} & \textbf{Harvest} & \textbf{Avg.} \\ \midrule
L2~\cite{r1}              & 4.10           & 8.41         & 8.89          & 8.39          & 14.65         & 14.92           & 18.50            & 19.80             & 25.60         & 30.62            & 15.39         \\
WG10~\cite{r35}            & 2.06          & 6.73         & 7.18          & 6.50           & 13.12         & 10.91           & 15.70            & 15.39            & 25.89        & 30.01            & 13.35         \\
AZ08~\cite{r26}            & 2.71          & 6.53         & 7.23          & 5.96          & 11.03         & 12.54           & 13.93           & 14.17            & 21.48        & 30.50             & 12.61         \\
GC10~\cite{r36}            & 3.21          & 8.22         & 8.53          & 6.62          & 7.90           & 14.85           & 14.22           & 19.07            & 9.55         & 27.84            & 12.00            \\
IA14~\cite{r37}            & 3.34          & 6.74         & 6.64          & 7.11          & 8.77          & 10.47           & 9.71            & 14.19            & 13.05        & 25.95            & 10.60          \\
ST14~\cite{r38}            & 1.74          & 6.12         & 6.51          & 6.12          & 8.78          & 10.60            & 10.09           & 13.63            & 13.93        & 25.44            & 10.30          \\
SPLINE-Net~\cite{r33}      & 4.51          & 6.49         & 8.29          & 5.28          & 10.89         & 10.36           & 9.62            & 15.50             & 7.44         & 17.93            & 9.63          \\
EW20~\cite{r39}            & 1.58          & 6.30          & 6.67          & 6.38          & 7.26          & 13.69           & 11.42           & 15.49            & 7.80          & 18.74            & 9.53          \\
DPSN~\cite{r23}            & 2.02          & 6.54         & 7.05          & 6.31          & 7.86          & 12.68           & 11.28           & 15.51            & 8.01         & 16.86            & 9.41          \\
CK18~\cite{r40}            & \underline{1.50}           & 5.74         & 6.24          & 4.97          & 8.64          & 8.86            & 10.00              & 11.44            & 11.33        & 21.90             & 9.06          \\
IRPS~\cite{r34}            & \textbf{1.47}          & 5.44         & 6.09          & 5.79          & 7.76          & 10.36           & 11.47           & 11.03            & 6.32         & 22.59            & 8.83          \\
LMPS~\cite{r32}            & 2.40           & 6.11         & 6.54          & 5.23          & 7.48          & 9.89            & 8.61            & 13.68            & 7.98         & 16.18            & 8.41          \\
PS-FCN~\cite{r5}          & 2.82          & 6.16         & 7.13          & 7.55          & 7.25          & 7.91            & 8.60             & 13.33            & 7.33         & 15.85            & 8.39          \\
Attention-PSN~\cite{r27}   & 2.93          & 6.14         & 6.92          & 4.86          & 6.97          & 7.75            & 8.42            & 12.90             & 6.86         & 15.44            & 7.92          \\
DR-PSN~\cite{r41}          & 2.27          & 5.42         & 7.08          & 5.46          & 7.21          & 7.84            & 8.49            & 12.74            & 7.01         & 15.40             & 7.90           \\
GPS-Net~\cite{r26}         & 2.92          & 5.42         & 6.04          & 5.07          & 7.01          & 7.77            & 9.00               & 13.58            & 6.14         & 15.14            & 7.81          \\
CHR-PSN~\cite{r8}         & 2.26          & 5.97         & 7.04          & 6.35          & 6.76          & \underline{7.15}            & 8.32            & 12.52            & 6.05         & 15.32            & 7.77          \\
CNN-PS~\cite{r6}          & 2.12          & \textbf{4.38}         & \underline{5.37}          & \underline{4.20}           & 6.38          & 8.07            & \underline{7.42}            & 12.12            & 7.92         & 13.83            & 7.18         \\
PS-FCN(Norm.)~\cite{r42}   & 2.67          & 4.76         & 6.17          & 7.72          & 7.15          & 7.53            & 7.84            & 10.92            & 6.72         & 12.39            & 7.39          \\
PX-Net~\cite{r7}          & 2.03          & \underline{4.39}         & \textbf{5.08}          & \textbf{4.13}          & \textbf{5.10}           & 7.61            & \textbf{6.90}             & 10.26            & \textbf{4.69}         & 13.10             & \textbf{6.33}          \\
NormAttention-PSN \cite{r43}   & 2.93          & 4.65         & 5.96          & 5.48          & 6.42          & \textbf{7.12}            & 7.49            & \underline{9.93}             & \underline{5.99}         & \underline{12.28}            & \underline{6.83}          \\
RMAFF-PSN(Ours) & 2.18          & 4.68         & 5.52          & 5.00             &  \underline{6.33}    & 7.71            & 8.26            & \textbf{9.46}    & 7.50          & \textbf{12.23}   & 6.89          \\ \bottomrule
\end{tabular}}
\noindent{\footnotesize{\textsuperscript{} The value represents the MAE of the estimated surface normals, where the \textbf{bold} numbers indicate the best results, while the  \underline{underlined} values represent the second-best performance.}}
 \label{table2}
\end{table}
\subsection*{Qualitative Comparison of Other Real-World Datasets\\}
To  validate the efficacy and generalizability of our proposed approach, we conducted qualitative experiments on three real datasets,namely, Apple\text and Gourd, DiLiGenT${10}^2$, and Simple PS data. Thanks to the richer features extracted by the RMAFF-PSN and the avoidance of over-smoothing in structurally complex regions, our method can recover clearer surface normals.
\par
\begin{figure}[htb]

    \includegraphics[width=12.5 cm]{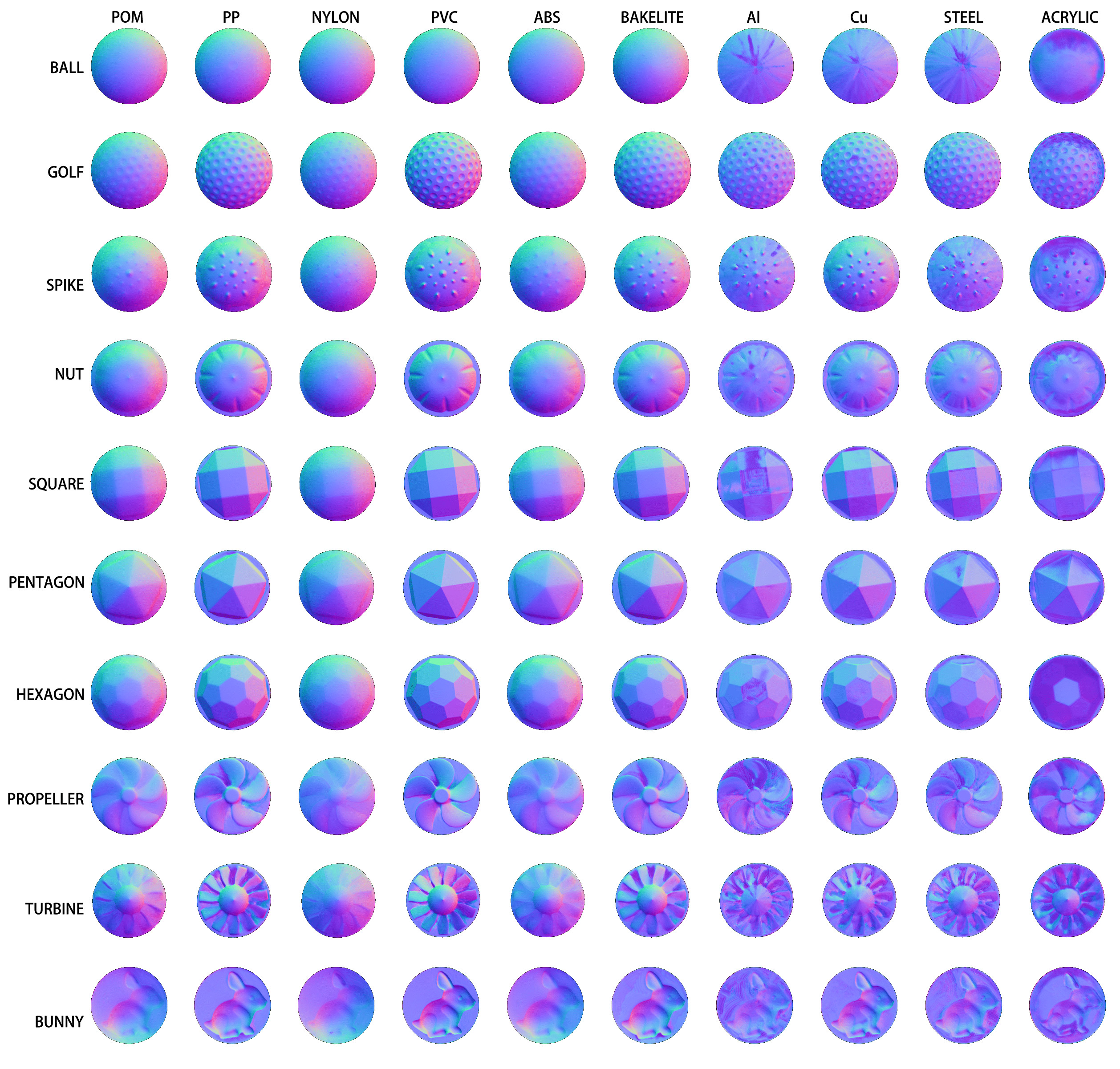}
    \caption{We present qualitative results for the DiLiGenT${10}^2$  dataset.\ From left to right, we demonstrate the robustness of our network for materials with varying isotropy and anisotropy groups, including challenge groups such as acrylic. We also show the effect of the network on objects with different surface structures and global illumination conditions, from top to bottom.}
    \label{fig:diligent100}
\end{figure}

The DiLiGenT${10}^2$ dataset provides a comprehensive evaluation of our network's prediction results for objects with diverse shapes and material groups. As shown in Figure~\ref{fig:diligent100}, our network's performance is limited when dealing with objects made of transparent acrylic. This is primarily due to the lack of such objects in the training dataset, resulting in a significant bias towards predicting pixel normals for this challenging material.

\par
\begin{figure}[htb]

    \includegraphics[width=12.5 cm]{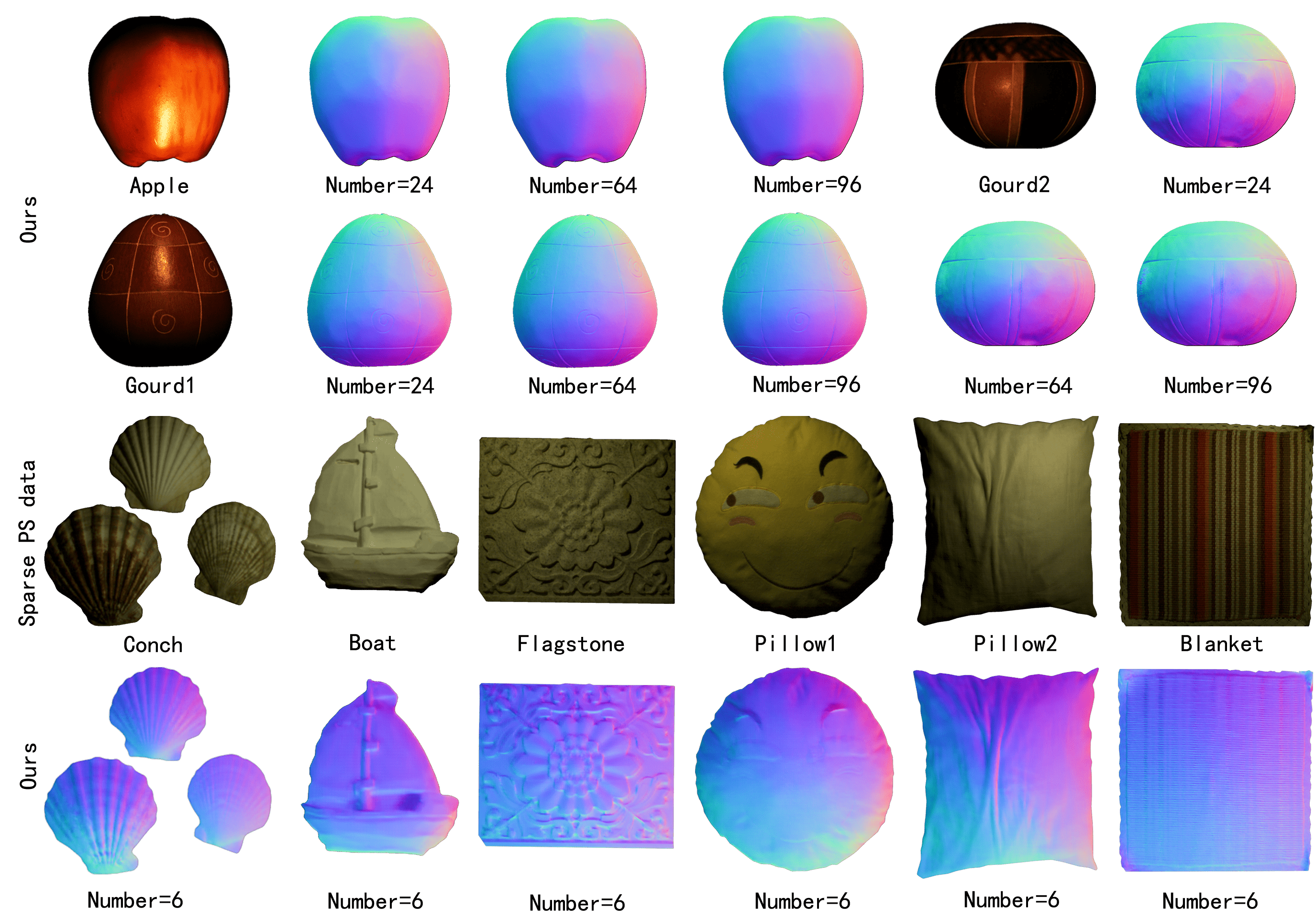}
    \caption{We present qualitative results of our RMAFF-PSN on the objects ``Apple'', “Gourd1” and “Gourd2” from the Apple and Gourd dataset, where 24, 64, and 96 represent the numbers of input images. We also shot the dataset under sparse light directions. As shown in the third and fourth rows, our method is able to reconstruct clear and normal directions of the objects with only six input images.}
    \label{fig:otherDatasets}
\end{figure}

To assess the generalization ability of our network model under sparse illumination, we acquired a new dataset using an industrial camera. This dataset,  includes objects with surfaces obscured by shadows (e.g.,``Conch'', ``Flagstone'') as well as challenging wool-like materials (e.g.,``Pillow1'', ``Blanket''). Figure  \ref{fig:otherDatasets} displays the normal prediction results for all six input images. Despite the sparse lighting conditions, our network accurately predicts the surface normals of the objects.

\subsection*{Discussion\\}
The network model proposed in this paper can promote the application of photometric stereoscopic technology in some 3D modeling fields requiring fine detail, such as industrial defect detection, film, computer-generated images, etc. In addition, the results of this paper show that enhancing and optimizing the retained normal-related channel information in the feature map and reducing the non-normal-related information (such as light intensity) are effective for the prediction of the normal of the complex structure region in the photometric stereoscopic task. Although we tested the resilience of our method under dense and sparse lighting conditions, we obtained fuzzy reconstruction results for some rare object surface materials, such as ``acrylic''. We infer that the reason for this result is that the object materials in our training dataset are single and lack some ``challenging'' materials. We will continue to investigate this phenomenon in our future work.

\par
\section*{Conclusions}
In this paper, we introduce a novel multi-scale feature fusion network that addresses the problem of a blurred normal reconstruction of ``difficult'' regions in the  calibrated photometric stereo problem with improved efficacy. Our approach leverages shallow-deep branching features, multi-level feature enhancement, and mixed attention-weighted optimization to achieve high-performance results. Using the DiLiGenT benchmark and additional real-world datasets, we demonstrate that our network generates accurate surface reconstructions, especially in regions with intricate structures.
\par
To validate the scalability and versatility of our proposed multi-scale feature fusion network, we conducted ablation experiments to showcase its efficacy in addressing the challenging problem of sparse photometric stereo. We believe that the robustness and scalability of our approach make it suitable for a wide range of real-world applications. In future work, we plan to upgrade existing shooting equipment and enhance the adaptability of our model by training it on more diverse and complex datasets, thereby expanding its capabilities and enabling it to handle a wider range of real-world textures and shapes.



\begin{thebibliography}{99}

\bibitem{r1}
Woodham, R.J.
\newblock Photometric method for determining surface orientation from multiple
  images.
\newblock {\em Opt. Eng.} {\bf 1980}, {\em 19},~139--144.

\bibitem{r2}
Sun, Z.L.; Lam, K.M.
\newblock Depth estimation of face images based on the constrained ICA model.
\newblock {\em IEEE Trans. Inf. Forensics Secur.} {\bf
  2011}, {\em 6},~360--370.

\bibitem{r3}
Nie, W.Z.; Liu, A.A.; Zhao, S.; Gao, Y.
\newblock Deep correlated joint network for 2-d image-based 3-d model
  retrieval.
\newblock {\em IEEE Trans. Cybern.} {\bf 2020}, \emph{52}, 1862--1871.

\bibitem{r4}
Jian, M.; Dong, J.; Gong, M.; Yu, H.; Nie, L.; Yin, Y.; Lam, K.M.
\newblock Learning the traditional art of Chinese calligraphy via
  three-dimensional reconstruction and assessment.
\newblock {\em IEEE Trans. Multimed.} {\bf 2019}, {\em 22},~970--979.

\bibitem{r5}
Chen, G.; Han, K.; Wong, K.Y.K.
\newblock PS-FCN: A flexible learning framework for photometric stereo.
\newblock In \emph{Proceedings of the European Conference on
  Computer Vision (ECCV),  2018}; Springer: Berlin, Germany, September, 2018; pp. 3--18. 

\bibitem{r6}
Ikehata, S.
\newblock CNN-PS: CNN-based photometric stereo for general non-convex surfaces.
\newblock In \emph{Proceedings of the European Conference on
  Computer Vision (ECCV),  2018}; Springer: Berlin, Germany, September, 2018; pp. 3--18.

\bibitem{r7}
Logothetis, F.; Budvytis, I.; Mecca, R.; Cipolla, R.
\newblock PX-NET: Simple and Efficient Pixel-Wise Training of Photometric Stereo Networks.
\newblock In \emph{Proceedings of the IEEE/CVF International Conference on Computer Vision (ICCV); 
}; ICCV Press : IEEE Computer Society, 2021; pp. 12757--12766.

\bibitem{r8}
Ju, Y.; Peng, Y.; Jian, M.; Gao, F.; Dong, J.
\newblock Learning conditional photometric stereo with high-resolution
  features.
\newblock {\em Comput. Vis. Media} {\bf 2022}, {\em 8},~105--118.

\bibitem{r9}
Liu, Y.; Ju, Y.; Jian, M.; Gao, F.; Rao, Y.; Hu, Y.; Dong, J.
\newblock A deep-shallow and global--local multi-feature fusion network for
  photometric stereo.
\newblock {\em Image Vis. Comput.} {\bf 2022}, {\em 118},~104368.

\bibitem{r10}
de~Santana~Correia, A.; Colombini, E.L.
\newblock Attention, please! A survey of neural attention models in deep
  learning.
\newblock {\em Artif. Intell. Rev.} {\bf 2022}, \emph{55}, 6037--6124.

\bibitem{r11}
Woo, S.; Park, J.; Lee, J.Y.; Kweon, I.S.
\newblock CBAM: Convolutional Block Attention Module.
\newblock In \emph{Proceedings of the European Conference on Computer Vision (ECCV)
2018}; Munich, Germany, September, 2018; pp. 3--19.

\bibitem{r12}
He, K.; Zhang, X.; Ren, S.; Sun, J.
\newblock Deep residual learning for image recognition.
\newblock In \emph{Proceedings of the IEEE Conference on Computer Vision and Pattern Recognition (CVPR)}; IEEE: Las Vegas, NV, USA, June, 2016; pp. 770--778.

\bibitem{r13}
Gao, S.H.; Cheng, M.M.; Zhao, K.; Zhang, X.Y.; Yang, M.H.; Torr, P.
\newblock Res2net: A new multi-scale backbone architecture.
\newblock {\em IEEE Trans. Pattern Anal. Mach. Intell.}
  {\bf 2019}, {\em 43},~652--662.

\bibitem{r14}
Zhu, H.; Li, P.; Xie, H.; Yan, X.; Liang, D.; Chen, D.; Wei, M.; Qin, J.
\newblock I can find you! Boundary-guided separated attention network for
  camouflaged object detection.
\newblock In \emph{Proceedings of the AAAI Conference on
  Artificial Intelligence,  2022}; AAAI Press: Palo Alto, CA, USA; Vancouver, Canada, February, 2022; Volume 36; pp. 3608--3616.

\bibitem{r15}
Santo, H.; Samejima, M.; Sugano, Y.; Shi, B.; Matsushita, Y.
\newblock Deep photometric stereo network.
\newblock In Proceedings of the IEEE International
  Conference on Computer Vision Workshops, Venice, Italy, 22--29 October 2017; pp. 501--509.

\bibitem{r16}
Zheng, Q.; Shi, B.; Pan, G.
\newblock Summary study of data-driven photometric stereo methods.
\newblock {\em Virtual Real. Intell. Hardw.} {\bf 2020}, {\em
  2},~213--221.

\bibitem{r17}
Ju, Y.; Lam, K.M.; Xie, W.; Zhou, H.; Dong, J.; Shi, B.
\newblock Deep Learning Methods for Calibrated Photometric Stereo and Beyond: A
  Survey.
\newblock {\em arXiv Preprint} {\bf 2022}, arXiv:2212.08414.

\bibitem{r18}
Yao, Z.; Li, K.; Fu, Y.; Hu, H.; Shi, B.
\newblock Gps-net: Graph-based photometric stereo network.
\newblock {\em Adv. Neural Inf. Process. Syst.} {\bf 2020},
  {\em 33},~10306--10316.

\bibitem{r19}
Ju, Y.; Lam, K.M.; Chen, Y.; Qi, L.; Dong, J.
\newblock Pay attention to devils: A photometric stereo network for better
  details.
\newblock In Proceedings of the Twenty-Ninth International
  Conference on International Joint Conferences on Artificial Intelligence,
  2021, Yokohama, Japan,  7--15 January 2021; pp. 694--700.

\bibitem{r20}
Jensen, H.W.; Marschner, S.R.; Levoy, M.; Hanrahan, P.
\newblock A practical model for subsurface light transport.
\newblock In Proceedings of the 28th Annual Conference on
  Computer Graphics and Interactive Techniques, Los Angeles, CA, USA, 12--17 August 2001; pp.~511--518.

\bibitem{r21}
Huang, G.; Liu, Z.; Van Der~Maaten, L.; Weinberger, K.Q.
\newblock Densely connected convolutional networks.
\newblock In Proceedings of the IEEE Conference on Computer
  Vision and Pattern Recognition, Honolulu, HI, USA, 21--26 July 2017; pp. 4700--4708.

\bibitem{r22}
Johnson, M.K.; Adelson, E.H.
\newblock Shape estimation in natural illumination.
\newblock In \emph{Proceedings of the CVPR 2011}; IEEE: Colorado Springs, CO, USA, June, 2011; pp. 2553--2560.

\bibitem{r23}
Wiles, O.; Zisserman, A.
\newblock SilNet: Single-and multi-view reconstruction by learning from
  silhouettes.
\newblock {\em arXiv Preprint} {\bf 2017}, arXiv:1711.07888.

\bibitem{r24}
Matusik, W.
\newblock A Data-Driven Reflectance Model.
\newblock Ph.D. Thesis, Massachusetts Institute of Technology, Cambridge, MA, USA, 2003.

\bibitem{r25}
Shi, B.; Wu, Z.; Mo, Z.; Duan, D.; Yeung, S.K.; Tan, P.
\newblock A benchmark dataset and evaluation for non-lambertian and
  uncalibrated photometric stereo.
\newblock In \emph{Proceedings of the IEEE Conference on Computer Vision and Pattern Recognition (CVPR)}; IEEE: Las Vegas, NV, USA, June, 2016; pp. 3707--3716.

\bibitem{r26}
Alldrin, N.; Zickler, T.; Kriegman, D.
\newblock Photometric stereo with non-parametric and spatially-varying
  reflectance. 
\newblock In Proceedings of the 2008 IEEE Conference on Computer Vision and
  Pattern Recognition, Anchorage, AK, USA , 23--28 June 2008; pp. 1--8.

\bibitem{r27}
Ren, J.; Wang, F.; Zhang, J.; Zheng, Q.; Ren, M.; Shi, B.
\newblock DiLiGenT102: A Photometric Stereo Benchmark Dataset With Controlled
  Shape and Material Variation.
\newblock In Proceedings of the IEEE/CVF Conference on
  Computer Vision and Pattern Recognition,  New Orleans, LA, USA, 18--24 June 2022; pp. 12581--12590.

\bibitem{r28}
Dai, Y.; Gieseke, F.; Oehmcke, S.; Wu, Y.; Barnard, K.
\newblock Attentional feature fusion.
\newblock In Proceedings of the IEEE/CVF Winter Conference
  on Applications of Computer Vision, Virtual Conference, 5--9 January 2021; pp. 3560--3569.

\bibitem{r29}
Sun, Y.; Wang, S.; Chen, C.; Xiang, T.Z.
\newblock Boundary-guided camouflaged object detection.
\newblock {\em arXiv Preprint} {\bf 2022}, arXiv:2207.00794.

\bibitem{r30}
Bolon-Canedo, V.; Remeseiro, B.
\newblock Feature selection in image analysis: a survey.
\newblock {\em Artif. Intell. Rev.} {\bf 2020}, {\em
  53},~2905--2931.

\bibitem{r31}
Kabir, H.; Garg, N.
\newblock Machine learning enabled orthogonal camera goniometry for accurate
  and robust contact angle measurements.
\newblock {\em Sci. Rep.} {\bf 2023}, {\em 13},~1497.

\bibitem{r32}
Li, J.; Robles-Kelly, A.; You, S.; Matsushita, Y.
\newblock Learning to minify photometric stereo.
\newblock In Proceedings of the IEEE/CVF Conference on
  Computer Vision and Pattern Recognition, Long Beach, CA, USA, 16--17 June 2019; pp. 7568--7576.

\bibitem{r33}
Zheng, Q.; Jia, Y.; Shi, B.; Jiang, X.; Duan, L.Y.; Kot, A.C.
\newblock SPLINE-Net: Sparse photometric stereo through lighting interpolation
  and normal estimation networks.
\newblock In Proceedings of the IEEE/CVF International
  Conference on Computer Vision,  Seoul, Republic of Korea, 27 October--2 November 2019; pp. 8549--8558.

\bibitem{r34}
Taniai, T.; Maehara, T.
\newblock Neural inverse rendering for general reflectance photometric stereo.
\newblock In Proceedings of the International Conference on Machine Learning.
  PMLR, Stockholm, Sweden, 10--15 July 2018; pp. 4857--4866.

\bibitem{r35}
Wu, L.; Ganesh, A.; Shi, B.; Matsushita, Y.; Wang, Y.; Ma, Y.
\newblock Robust photometric stereo via low-rank matrix completion and
  recovery.
\newblock In \emph{Proceedings of the Asian Conference on Computer Vision}; Springer: Berlin, Germany,
  2011; pp. 703--717.

\bibitem{r36}
Goldman, D.B.; Curless, B.; Hertzmann, A.; Seitz, S.M.
\newblock Shape and spatially-varying brdfs from photometric stereo.
\newblock {\em IEEE Trans. Pattern Anal. Mach. Intell.}
  {\bf 2009}, {\em 32},~1060--1071.

\bibitem{r37}
Ikehata, S.; Aizawa, K.
\newblock Photometric stereo using constrained bivariate regression for general
  isotropic surfaces.
\newblock In Proceedings of the IEEE Conference on Computer
  Vision and Pattern Recognition, Columbus, OH, USA,  23--28 June 2014; pp. 2179--2186.

\bibitem{r38}
Shi, B.; Tan, P.; Matsushita, Y.; Ikeuchi, K.
\newblock Bi-polynomial modeling of low-frequency reflectances.
\newblock {\em IEEE Trans. Pattern Anal. Mach. Intell.}
  {\bf 2013}, {\em 36},~1078--1091.

\bibitem{r39}
Enomoto, K.; Waechter, M.; Kutulakos, K.N.; Matsushita, Y.
\newblock Photometric stereo via discrete hypothesis-and-test search.
\newblock In Proceedings of the IEEE/CVF Conference on
  Computer Vision and Pattern Recognition, Seattle, WA, USA, 13--19 June 2020; pp. 2311--2319.

\bibitem{r40}
Simchony, T.; Chellappa, R.; Shao, M.
\newblock Direct analytical methods for solving Poisson equations in computer
  vision problems.
\newblock {\em IEEE Trans. Pattern Anal. Mach. Intell.}
  {\bf 1990}, {\em 12},~435--446.

\bibitem{r41}
Ju, Y.; Dong, J.; Chen, S.
\newblock Recovering surface normal and arbitrary images: A dual regression
  network for photometric stereo.
\newblock {\em IEEE Trans. Image Process.} {\bf 2021}, {\em
  30},~3676--3690.

\bibitem{r42}
Chen, G.; Han, K.; Shi, B.; Matsushita, Y.; Wong, K.Y.K.
\newblock Deep photometric stereo for non-lambertian surfaces.
\newblock {\em IEEE Trans. Pattern Anal. Mach. Intell.}
  {\bf 2020}, {\em 44},~129--142.

\bibitem{r43}
Ju, Y.; Shi, B.; Jian, M.; Qi, L.; Dong, J.; Lam, K.M.
\newblock Normattention-psn: A high-frequency region enhanced photometric
  stereo network with normalized attention.
\newblock {\em Int. J. Comput. Vis.} {\bf 2022}, {\em
  130},~3014--3034.

\end{thebibliography}
\end{document}